\documentclass[runningheads]{llncs}
\usepackage[T1]{fontenc}
\usepackage{graphicx}
\usepackage{booktabs}
\usepackage{multirow}
\usepackage{array}

\usepackage{textgreek}

\newcommand{\reffigure}[1]{\figurename~\ref{#1}}

\newcommand{\reftable}[1]{Table~\ref{#1}}

\begin{document}
\title{For How Long Should We Be Punching?\\ Learning Action Duration in Fighting Games}
\titlerunning{Learning Action Duration in Fighting Games}
%
\author{Hoang Hai Nguyen \and
Kurt Driessens\orcidID{0000-0001-7871-2495} \and \\
Dennis J.N.J. Soemers\orcidID{0000-0003-3241-8957}}
\authorrunning{H. Nguyen et al.}
%
\institute{Department of Advanced Computing Sciences, Maastricht University,\\ Maastricht, the Netherlands \\
\email{hh-nguyen@student.maastrichtuniversity.nl} \\
\email{\{kurt.driessens, dennis.soemers\}@maastrichtuniversity.nl}}
\maketitle              
\begin{abstract}
Fighting games such as \textit{Street Fighter II} present unique challenges to reinforcement learning (RL) agents due to their fast-paced, real-time nature. In most RL frameworks, agents are hard-coded to make decisions at a fixed interval, typically every frame or every $N$ frames. Although this design ensures timely responses, it restricts the agent’s ability to adjust its reaction timing. Acting every frame grants frame-perfect reflexes, which are unrealistic compared to human players, whereas longer fixed intervals reduce computational cost but hinder responsiveness. We consider an alternative decision-making framework in which the agent learns not only \textit{what} action to take but also \textit{for how long} to execute it. By jointly predicting both action and duration, the agent can dynamically adapt its responsiveness to different situations in the game. We implement this method using the open-source \textit{FightLadder} environment with agents trained against scripted built-in bots, systematically testing different frame skip configurations to analyze their influence on performance, responsiveness, and learned behavior. Experiments show that learned timing can match the performance of well-chosen fixed frame skips and encourages repeatable action patterns, but does not ensure robustness on its own. In most cases, we see agents performing best with consistently high frame skip values (i.e., low responsiveness). This strategy makes it easier to learn exploitative strategies where the same action is repeated over and over, which the scripted bots appear to be susceptible to.

\end{abstract}

\section{Introduction}
Reinforcement learning (RL) has achieved impressive results in many environments, such as Atari games \cite{Mnih_2015_DQN}, and complex board games like Go where approaches like AlphaGo \cite{silver2016mastering} and AlphaZero \cite{silver2018general} achieved breakthrough results. However, fast-paced games remain a difficult challenge. These games require rapid decision-making, continuous adaptation, and careful timing of actions. 

Timing is a crucial aspect in fighting games, where subtle differences in action timings can lead to wildly varying results. 
In RL for video games, it is common practice to have agents select actions every $N$ frames for some---typically predetermined---positive integer $N$. When $N > 1$, which is typically the case, this is referred to as frame skipping. Intermediate frames are skipped from the perspective of the agent, which only observes one in every $N$ frames.
Fixed frame-skipping introduces two potential limitations for RL agents. If the agent makes decisions every single frame, it gains an unrealistically fast reaction time that humans cannot match. The increased number of decisions per game also imposes higher computational costs, and complicates the credit assignment \cite{McGovern_1998_MacroActions} and exploration \cite{Dabney_2021_TemporallyExtended} problems of RL \cite{Sutton_2018_RL,kalyanakrishnan2021analysis} due to the increased number of decisions that might be responsible for observed performance. However, if the gap between the decision points is too large, the agent becomes slow and potentially unresponsive at crucial moments.

We augment the action space in \textit{Street Fighter II - Special Champion Edition}, using the FightLadder framework \cite{Li_2024_FightLadder} to include a choice as to how many frames to skip. This enables RL agents to learn to autonomously decide how many frames to skip for each action in a state-dependent manner. We trained agents using proximal policy optimization (PPO) \cite{schulman2017proximal}, training and evaluating against a variety of built-in scripted bots. The key outcomes from our experiments are that agents can learn to effectively select well-performing frame skip values, but they tend to consistently select rather high frame skip values, with seemingly little to no regard for game state. When using a fixed, non-adaptive frame skip, similarly high values also work well. Built-in scripted bots in this game turn out to be susceptible to exploitative strategies in which the same action is repeated over and over, which in turn is easier to learn for RL when operating at a coarse time resolution.

\section{Background}

Reinforcement learning (RL) provides a framework for sequential decision making, where an agent interacts with an environment over discrete time steps. This process is commonly described as a Markov Decision Process (MDP), defined by a tuple \((\mathcal{S}, \mathcal{A}, f, r, \gamma)\), where \(\mathcal{S}\) is a state space, \(\mathcal{A}\) is an action space, \(f\) is a transition function, \(r\) is a reward function, and \(\gamma\) is a discount factor \cite{Sutton_2018_RL}. At each time step \(t\), the agent observes a state, selects an action, and then transitions to a new state and receives a reward based on the state reached. The objective is to learn a policy that maximizes the long-term sum of discounted rewards. 

\textit{Street Fighter II - Special Champion Edition} is a fast-paced 2D fighting game where player inputs are processed in discrete frames. The game runs at 60 frames per second. The game environment advances in fixed discrete time steps (frames), and at each step, the game computes the player's health, movement, and visual frame to show to the player. The agent can only observe the pixel-based visual state representation, not any internal raw representation. We construct the observation by stacking 12 past game frames sampled with an observation stride of 8, forming a tensor input to the policy network and critic of PPO. Stacking frames is a common technique in RL, as it enables the agent to understand the temporal dynamics (e.g., infer directions and velocities of moving objects). The reward function provided by the Fightladder framework \cite{Li_2024_FightLadder} is designed to encourage effective fighting behavior. It includes rewarding the agent for causing damage, penalizing the agent for taking damage, and a win/lose bonus at the end. 

FightLadder provides an action space aligned with the original game's control scheme for human players. The human action space is mapped into a transformed action space, consisting of two components: a motion action set $\{$defense, forward, jump, crouch, back flip, front flip, offensive crouch, defensive crouch$\}$ and an attack action set $\{$light punch, medium punch, heavy punch$\}$. Combining the motion and attack sets leads to a joint \textit{MultiDiscrete} action space with 48 possible action combinations. Moreover, a hardcoded special moves list is included in the action space so that the agent can directly access combos as single actions. 

\section{Learning Action Duration}

\begin{figure}[t]
    \centering
    \includegraphics[width=\linewidth]{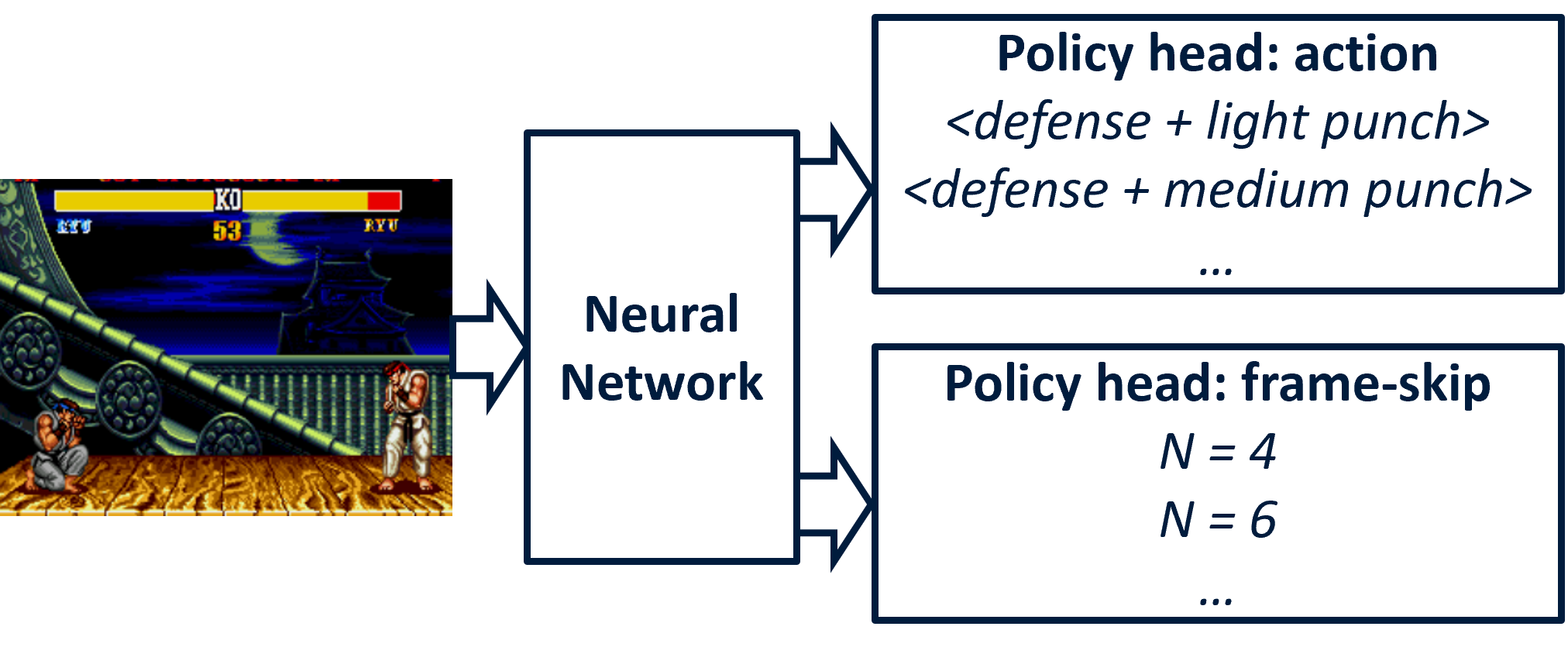}
    \caption{Policy architecture with separate heads. There is one policy head for combinations of movement and attack actions, and another for the selection of frame-skip.}
    \label{fig:sep}
\end{figure}

We use the standard PPO \cite{schulman2017proximal} training setup, with the standard convolutional neural network architecture and reward shaping, as implemented in FightLadder \cite{Li_2024_FightLadder}. The default FightLadder implementation uses a fixed frame-skip of $N = 8$, i.e., agents observe the game state and take decisions every 8 frames. During the skipped frames, the agent continues executing or repeating the action (i.e., specific type of movement and/or attack) that it selected for that decision point. We evaluate three variations on this theme: (i) using a fixed frame-skip for a variety of different values for $N$ (ranging from 4 to 60), (ii) sampling the number of frames to be skipped uniformly at random per decision from a predetermined range of possible values, and (iii) augmenting the action space of the RL agent to also output the desired frame-skip value for that decision.

\begin{figure}[t]
    \centering
    \includegraphics[width=\linewidth]{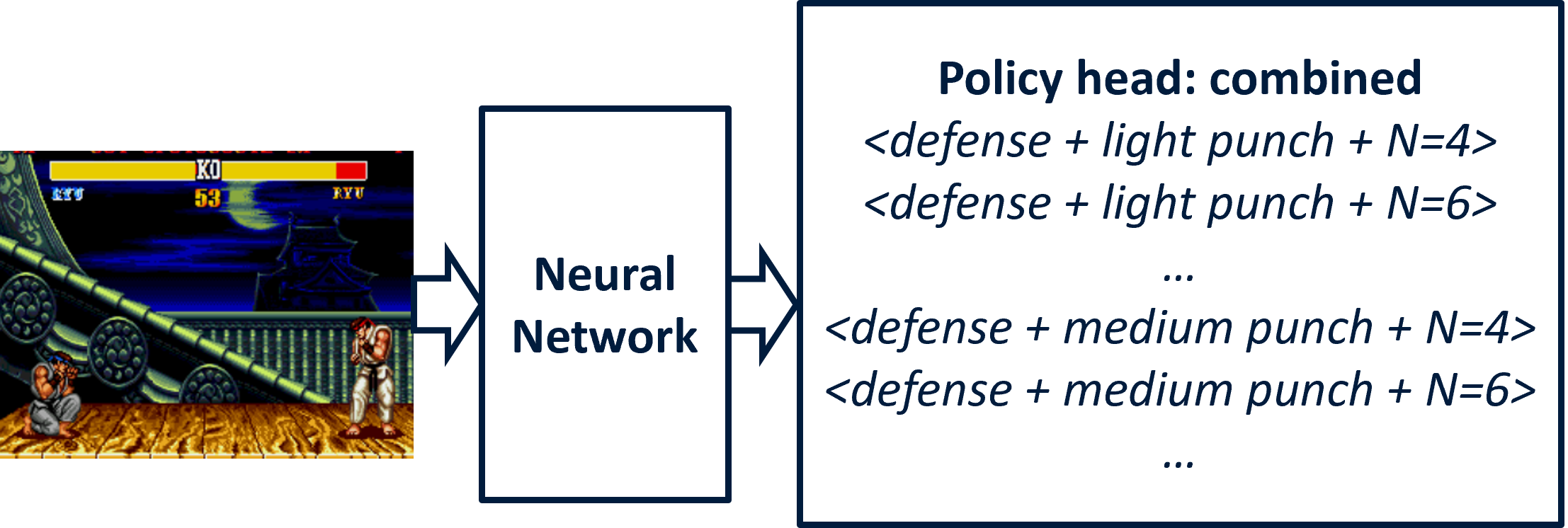}
    \caption{Policy architecture with combined head. Every action is a combination of movement, attack type, and frame-skip value.}
    \label{fig:comb}
\end{figure}

For the adaptive agent that learns to autonomously select its frame-skip value for each decision point, we furthermore consider two different ways to model the action space. In the \textit{separated} agent (see \reffigure{fig:sep}), the policy network has two independent output heads: one for the standard action space, and one for the frame-skip value. Actions are sampled from each head independently. In the \textit{combined} agent (see \reffigure{fig:comb}), the policy network only has a single output head for a joint action space, which is the product of the standard action space and the frame-skip choice. In either case, selecting a frame-skip value is modelled as a choice from a discrete action space (a limited number of frame-skip values).

\section{Experiments}

Throughout all experiments in this paper, we view the game as a single-agent game (from the RL perspective), training an agent to play the character \textit{Ryu}. Training is done against one or several different characters controlled by built-in bots from the game (which may be viewed as being a ``part of the environment'' from the RL perspective, as the opponents follow stationary, scripted policies).

\subsection{Setup of Experiments}

Unless stated otherwise, training parameters and hyperparameters were set as follows. Training proceeded over a total of 10 million timesteps. Experience was collected from 8 environments running in parallel. The learning rate was linearly decayed from 2.5e-4 to 2.5e-6. The clipping range was linearly decayed from 0.15 to 0.025. We used a batch size of 1024, training over 20 epochs. The coefficient for the entropy regularization term of the loss function was set to 0.01. A discount factor of 0.94 was used to discount rewards over time. Each parallel environment collected 512 steps worth of experience before each update. The dense and aggressive coefficients---two coefficients that are used in the FightLadder reward function---are annealed from 3.0 to 1.0, and from 1.0 to 0.0, respectively. These settings are all defaults in FightLadder.

We evaluate the performance of agents trained using different action spaces and frame-skipping strategies, described as follows. \textit{Fixed (N)} uses a predetermined, fixed frame-skip value of $N \in \{ 4, 8, 16, 60 \}$, where $N = 4$ is a commonly used value in many different games and environments throughout the deep RL literature popularized by the seminal work on Deep $Q$-networks by Mnih et al. \cite{Mnih_2015_DQN}, $N = 8$ is the default value used in FightLadder, and the other values are comparatively high extremes. \textit{Random (4-8)} and \textit{Random (4-16)} select, at each decision point, a frame-skip value uniquely at random from $\{ 4, 5, 6, 7, 8 \}$ or $\{ 4, 6, 8, 10, 12, 14, 16 \}$, respectively. \textit{Separated (4-8)}, \textit{Separated (4-16)}, and \textit{Separated (4-16,32)} use the separated policy heads (\reffigure{fig:sep}) for policy-driven selection of frame-skip values from $\{ 4, 5, 6, 7, 8 \}$, $\{ 4, 6, 8, 10, 12, 14, 16 \}$, or $\{ 4, 6, 8, 12, 14, 16, 32 \}$, respectively. Finally, \textit{Combined (4-8)} uses the combined policy head (\reffigure{fig:comb}) for policy-driven selection of frame-skip values from $\{ 4, 5, 6, 7, 8 \}$. Larger sets of frame-skip values for the combined architecture were not evaluated, as its action space scales poorly (multiplicatively with the size of the game's original action space) with increases in size of the set of values, and already with the smallest set this agent was found not to outperform the separated architecture in preliminary experiments.

Performance of trained agents is evaluated from evaluation games played post-training. In evaluation games, trained agents select the action(s) with the maximum probability output by the policy in $99\%$ of the decision points (focusing more on exploitation than exploration in post-training evaluation games), and select action(s) by sampling from the policy output in the remaining $1\%$ of decision points (to add more variety to evaluation games).

\subsection{Results}

\begin{table}[t]
\renewcommand{\arraystretch}{1.1}
\caption{$95\%$ Agresti-Coull confidence intervals for win percentages of trained agents against Ryu (built-in AI, cycling through star levels 1--8), over 100 evaluation games.}
\label{tab:ryu-winrates}
\centering
\begin{tabular}{@{}lr@{}}
\toprule
\textbf{Frame-skip strategy} \qquad \qquad & \textbf{Win percentage against Ryu} \\
\midrule
Fixed (4) & 74\% $\pm$ 8.5\% \\
Fixed (8) & 89\% $\pm$ 6.4\% \\
Fixed (16) & 100\% $\pm$ 2.6\% \\
Fixed (60) & 100\% $\pm$ 2.6\% \\
Random (4–8) & 79\% $\pm$ 8.0\% \\
Random (4–16) & 80\% $\pm$ 7.9\% \\ 
Separated (4–8) & 77\% $\pm$ 8.2\% \\
Separated (4–16) & 83\% $\pm$ 7.4\% \\
Separated (4-16,32) & 100\% $\pm$ 2.6\% \\ 
Combined (4–8) & 69\% $\pm$ 9.0\% \\
\bottomrule
\end{tabular}
\end{table}

\begin{figure}[t]
    \centering
    \includegraphics[width=\linewidth]{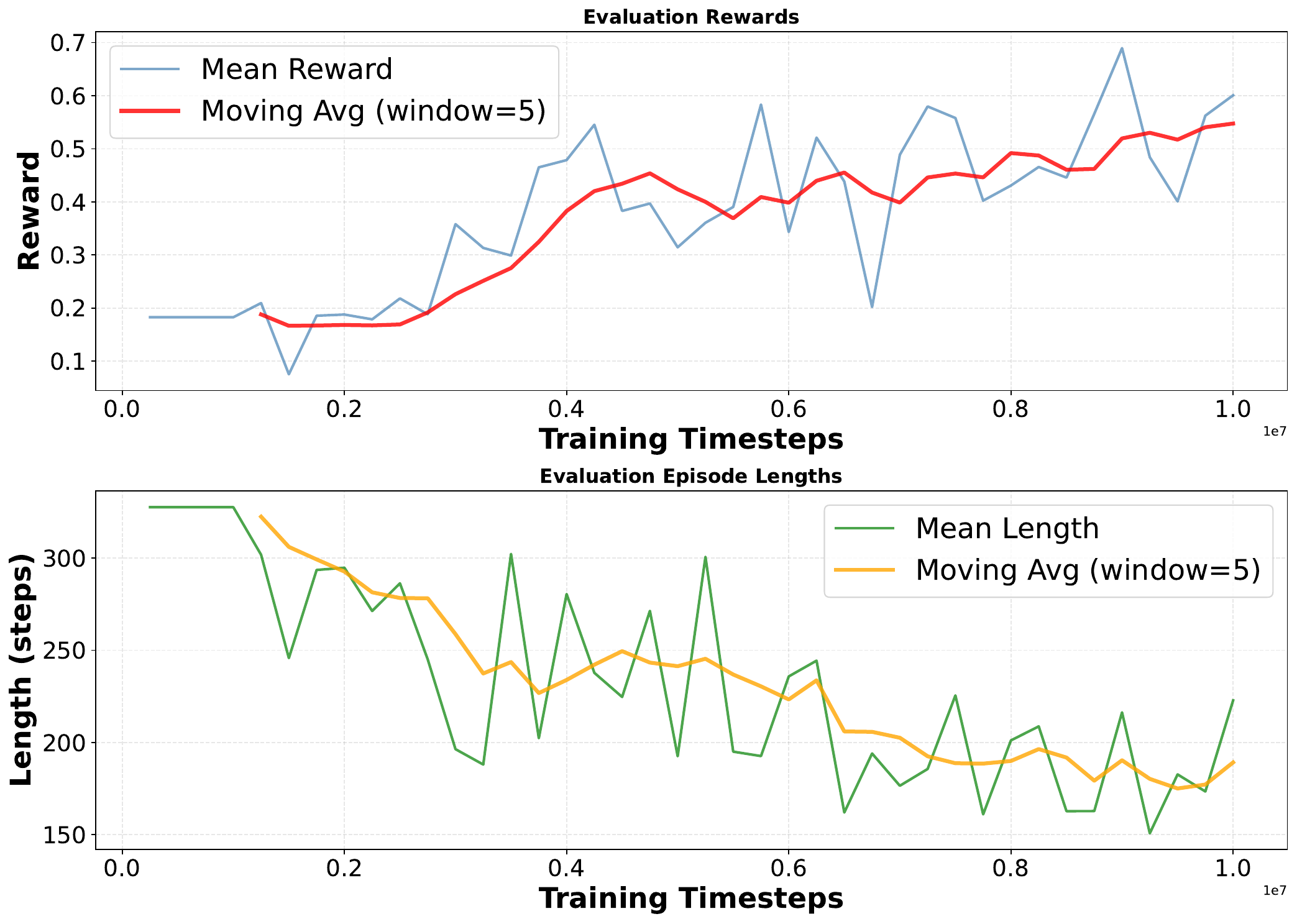}
    \caption{Average episodic reward (top) and episode length (bottom) for the \textit{Separated (4-16)} training run, as functions of number of training steps.}
    \label{fig:evaluation}
\end{figure}

\begin{figure}[t]
    \centering
    \includegraphics[width=\linewidth]{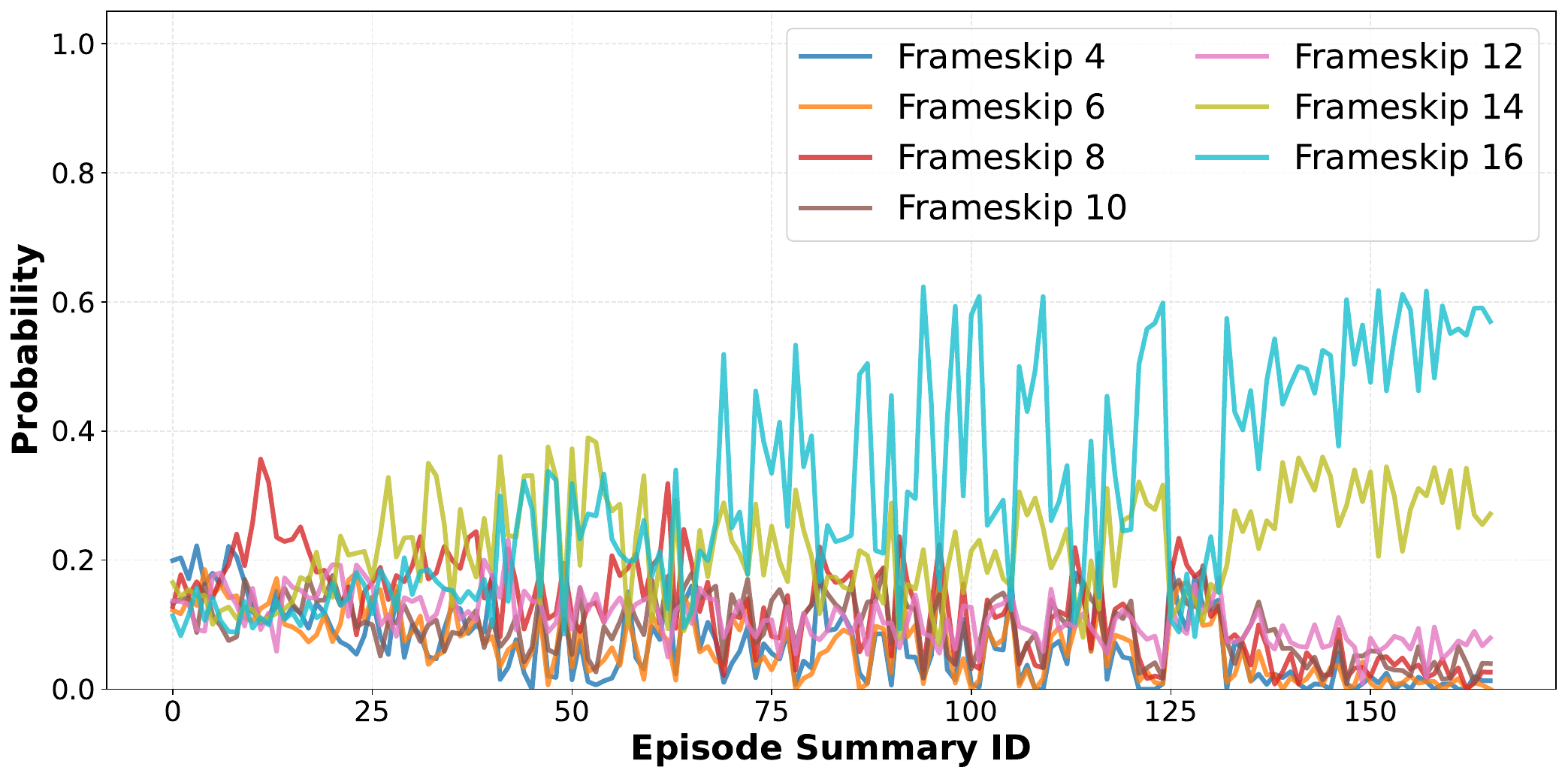}
    \caption{Distribution of frame-skip choices as function of the number of training episodes, for the \textit{Separated (4-16)} training run.}
    \label{fig:overall}
\end{figure}

\begin{figure}[t!]
    \centering
    \includegraphics[width=\linewidth]{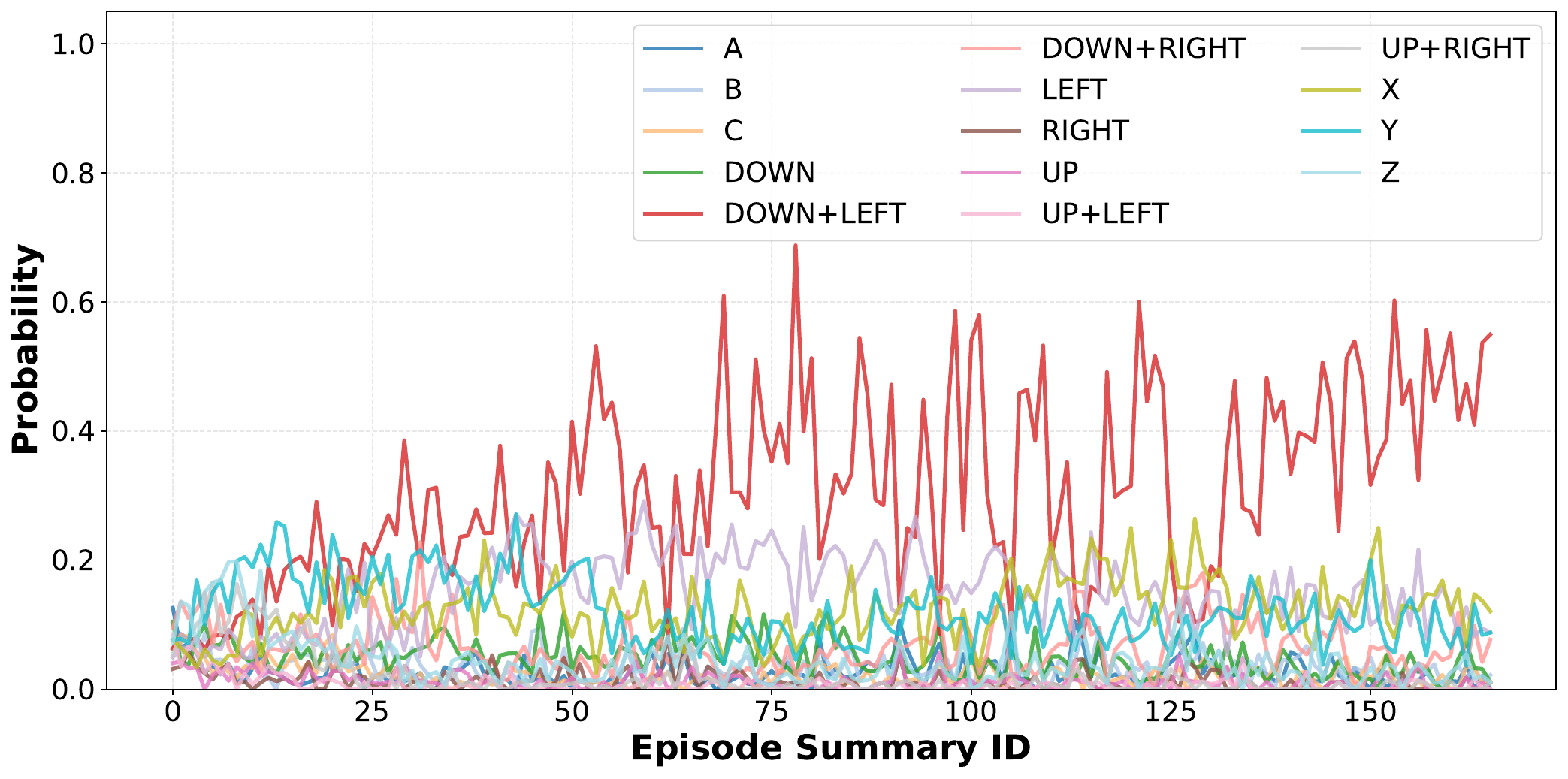}
    \caption{Distribution of button choices as function of the number of training episodes, for the \textit{Separated (4-16)} training run. Note that some actions (e.g., \textit{defense + light punch}) correspond to multiple simultaneous button presses.}
    \label{fig:button}
\end{figure}

\begin{figure}[t] 
    \centering
    \includegraphics[width=\linewidth]{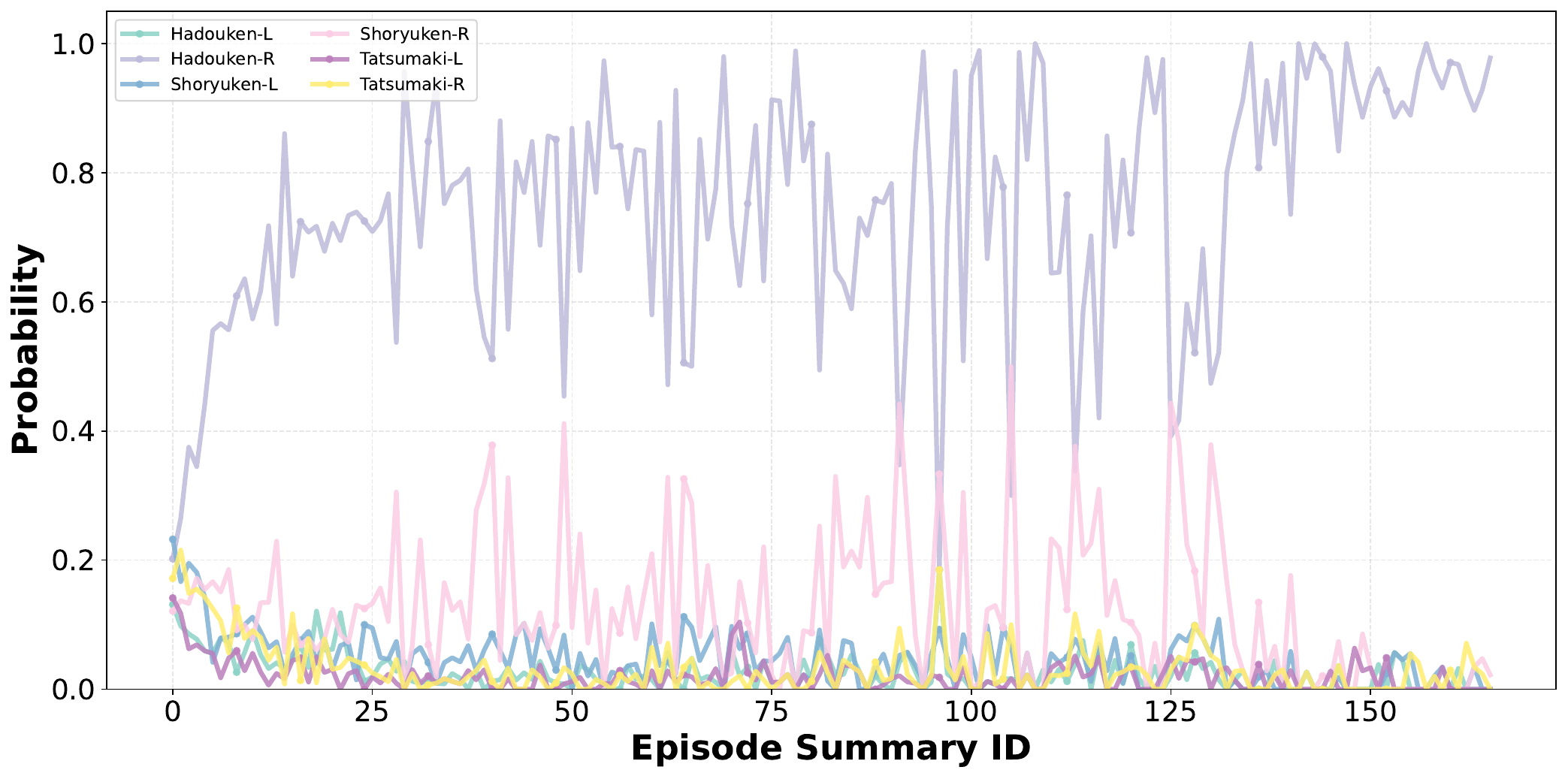}
    \caption{Distribution of combo choices as function of the number of training episodes, for the \textit{Separated (4-16)} training run.}
    \label{fig:combo}
\end{figure}

\reftable{tab:ryu-winrates} presents $95\%$ Agresti-Coull confidence intervals for the win percentages of trained agents using various frame-skip strategies against the built-in AI playing Ryu (i.e., the same character controlled by our trained agents), over a total of 100 evaluation games. The start level of the opponent (which serves as an indicator of difficulty level for human players) is varied from level 1 to 8 throughout these 100 evaluation games. The star level of the opponent was similarly varied during training (uniformly, not forming e.g. a curriculum of increasing difficulty level during training).

Zooming in on the \textit{Separated (4-16)} agent's training run as a case study, \reffigure{fig:evaluation} depicts how mean episodic rewards and episode lengths (measured in number of frames per episode) progress over training time. Similarly, \reffigure{fig:overall} depicts how the way in which the policy distributes probability mass over the different frame-skip choices changes as training progresses. Every data point in the figure averages the probability distributions output by the policy for all states encountered in the respective training episode. \reffigure{fig:button} and \reffigure{fig:combo} depict the progression of distributions for individual buttons and combos, respectively.

\reftable{tab:unseen-opponents} lists win percentages of trained agents against characters that were not seen during training (i.e., characters other than Ryu). Each of these is associated with a single level, serving as an indicator of difficulty. Considering the large number of opponents to evaluate against, we omitted the \textit{Combined} strategy from further consideration, and only ran 20 evaluation games per matchup for the remaining strategies. This means that confidence intervals (not shown) are not tight. The results are merely meant to give an impression of ability to generalize to new opponents, rather than establishing precise measures of win percentage.

\begin{table}[t]
\renewcommand{\arraystretch}{1.1}
\caption{Win percentages against opponents not seen during training, over 20 evaluation games per matchup. Opponent characters (listed in increasing order of difficulty level): \textbf{G = Guile (lvl1)}, \textbf{K = Ken (lvl2)}, \textbf{C = Chun-Li (lvl3)}, \textbf{Z = Zangief (lvl5)}, \textbf{D = Dhalsim (lvl6)}, \textbf{E = E. Honda (lvl9)}, \textbf{Bl = Blanka (lvl10)}, \textbf{Ba = Balrog (lvl11)}, \textbf{V = Vega (lvl13)}, \textbf{S = Sagat (lvl14)}, \textbf{Bi = Bison (lvl15)}.}
\label{tab:unseen-opponents}
\centering
\begin{tabular}{@{}lrrrrrrrrrrr@{}}
\toprule
\multirow{2}{*}[-0.3em]{\parbox{2cm}{\textbf{Frame-skip Strategy}}} & \multicolumn{11}{c}{\textbf{Opponent Character}} \\
\cmidrule(lr){2-12}
& \phantom{~}\phantom{~}\phantom{~}\phantom{~}\textbf{G} & \phantom{~}\phantom{~}\phantom{~}\phantom{~}\textbf{K} & \phantom{~}\phantom{~}\phantom{~}\phantom{~}\textbf{C} &
\phantom{~}\phantom{~}\phantom{~}\phantom{~}\textbf{Z} & \phantom{~}\phantom{~}\phantom{~}\phantom{~}\textbf{D} & \phantom{~}\phantom{~}\phantom{~}\phantom{~}\textbf{E} &
\phantom{~}\phantom{~}\phantom{~}\phantom{~}\textbf{Bl.} & \phantom{~}\phantom{~}\phantom{~}\phantom{~}\textbf{Ba} & \phantom{~}\phantom{~}\phantom{~}\phantom{~}\textbf{V} &
\phantom{~}\phantom{~}\phantom{~}\phantom{~}\textbf{S} & \phantom{~}\phantom{~}\phantom{~}\phantom{~}\textbf{Bi} \\
\midrule
Fixed (4) & 0\% & 85\% & 30\% & 20\% & 5\% & 5\% & 0\% & 5\% & 5\% & 0\% & 0\% \\ 
Fixed (8) & 10\% & 5\% & 15\% & 90\% & 5\% & 0\% & 0\% & 0\% & 0\% & 0\% & 0\% \\ 
Fixed (16) & 5\% & 5\% & 0\% & 5\% & 0\% & 0\% & 0\% & 0\% & 0\% & 0\% & 60\% \\ 
Fixed (60) & 10\% & 70\% & 15\% & 0\% & 15\% & 0\% & 0\% & 0\% & 0\% & 0\% & 0\% \\ 
Random (4-8) & 30\% & 5\% & 15\% & 25\% & 0\% & 0\% & 0\% & 0\% & 0\% & 0\% & 0\% \\ 
Random (4-16) & 100\% & 45\% & 50\% & 30\% & 25\% & 5\% & 15\% & 30\% & 30\% & 20\% & 10\% \\  
Separated (4-8) & 85\% & 80\% & 25\% & 10\% & 40\% & 0\% & 0\% & 0\% & 0\% & 0\% & 0\% \\ 
Separated (4-16) & 100\% & 10\% & 60\% & 65\% & 15\% & 0\% & 5\% & 25\% & 0\% & 0\% & 0\% \\ 
Separated (4-16,32)\phantom{~} & 100\% & 0\% & 90\% & 5\% & 25\% & 0\% & 5\% & 5\% & 10\% & 0\% & 0\% \\ 
\bottomrule
\end{tabular}
\end{table}

\reftable{tab:finetuning-results} presents win percentages, measured over 20 evaluation games per matchup, of the \textit{Separated (4-16)} agent being finetuned against new opponent characters after the initial training run against Ryu, using two different finetuning strategies. In the \textit{Single Finetuning} strategy, the policy trained against Ryu is finetuned separately against each individual opponent character. In the \textit{Sequential Finetuning} strategy, a single network (taking the policy trained against Ryu as a starting point) is finetuned sequentially against each of the new characters in increasing order of difficulty level. Finetuning runs start with a lower learning rate (decaying from 5.0e-5 to 2.5e-6) and clipping range (decaying from 0.075 to 0.025), and use only 2 million steps instead of 10 million. These changes to the default parameters are used because the finetuning runs start from networks that have already been trained (against Ryu), as opposed to randomly initialized networks.

\begin{table}[t] 
\centering
\caption{Win percentages over 20 evaluation games of the \textit{Separated (4-16)} agent finetuned using two different finetuning strategies after original training against Ryu.}
\label{tab:finetuning-results}
\begin{tabular}{@{}lrr@{}}
\toprule
\textbf{Opponent character (level)} \quad \quad &
\textbf{Single Finetuning} & 
\quad \textbf{Sequential Finetuning} \\
\midrule
Guile (lvl1) & 100\% & 100\%  \\
Ken (lvl2) & 5\% & 35\% \\
Chun-Li (lvl3) & 100\% & 40\% \\
Zangief (lvl5) & 90\% & 100\% \\
Dhalsam (lvl6) & 35\% & 75\% \\
E. Honda (lvl9) & 80\% & 100\% \\  
Blanka (lvl10) & 100\% & 85\% \\ 
Balrog (lvl11) & 95\% & 30\% \\ 
Vega (lvl13) & 20\% & 95\% \\  
Sagat (lvl14) & 5\% & 15\% \\  
Bison (lvl15) & 10\% & 15\% \\  
\bottomrule
\end{tabular}
\end{table}

\begin{figure}[t]
    \centering
    \includegraphics[width=\linewidth]{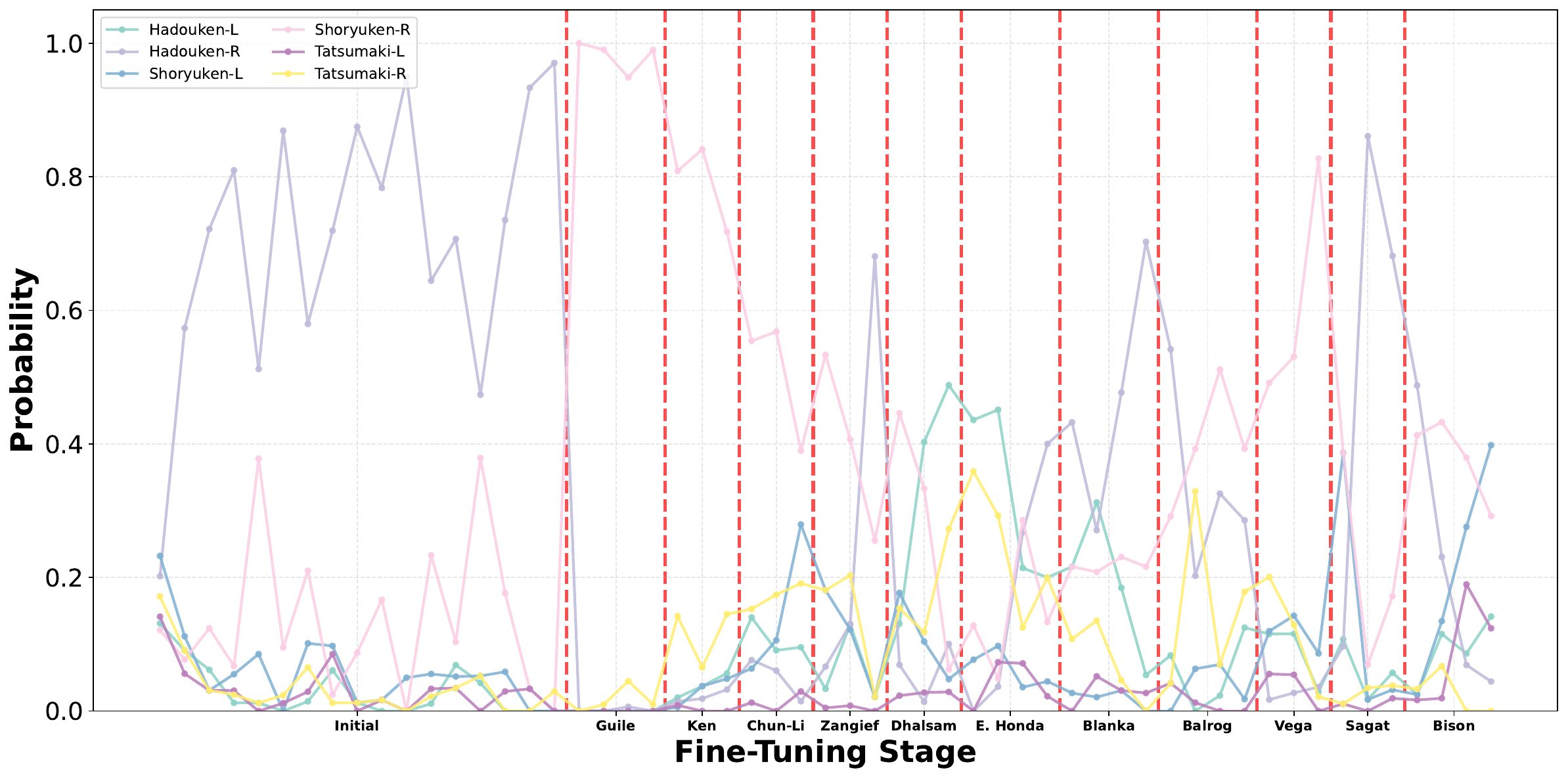}
    \caption{Distribution of combo choices as function of training time (measured in number of training episodes), for the \textit{Sequential Finetuning} training run (including the original \textit{Separated (4-16)} training run against Ryu in the first part). The vertical dashed lines indicate points in time where we switch to a new opponent character for training.}
    \label{fig:combo_seq}
\end{figure}

\reffigure{fig:combo_seq} depicts the progression of policy probability distributions for combo moves over training time in the \textit{Sequential Finetuning} setting. The vertical dashed lines represent the points in training time where we switch to a new training opponent.

\subsection{Discussion}

The results in \reftable{tab:ryu-winrates} suggest that agents can learn effective frame-skip values, in the sense that agents that need to select frame-skip values themselves instead of being assigned predetermined values have competitive levels of performance (they do not underperform). \reffigure{fig:evaluation} provides additional evidence that learning is stable and collected rewards (against the opponent character being trained against) continue to improve. However, the ability for an agent to select its own frame-skip value as a function of game state does not appear to provide any benefits over a well-tuned fixed frame-skip value in the tested training setup (this might still change with, e.g., longer training durations or different training partners). Furthermore, the best fixed frame-skip values turn out to be surprisingly high; whereas we may initially expect high responsiveness (low frame-skip) to be important in fast-paced fighting games, it turns out that even an agent with a frame-skip value of 60 (only making a new decision every full second) obtains a 100\% winrate versus the character we train against.

When an agent is allowed to select its own frame-skip values as a function of state, we observe that it also tends to learn to consistently pick relatively high frame-skip values (\reffigure{fig:overall}). Additional analysis (plots omitted due to space constraints) revealed no apparent relation to either player hit points (where we might have expected a player to prioritise responsiveness when at risk, i.e. at low hit points) or distance to opponent (where we might have expected a player to prioritise responsiveness when close to the opponent); the agent would continue to select high frame-skip values regardless of hit points or distance to opponent. A more detailed inspection of the learned behavior (\reffigure{fig:button} and~\ref{fig:combo}) reveals that the agent also tends to keep repeating the same actions in the game, with little variety in types of movements and attacks. This suggests that the agent learns strategies that could be considered exploitative: arguably uninteresting strategies that happen to work against the particular built-in AI we train against, but unlikely to generalize to other opponents. Such strategies are easier to learn with high frame-skip values, as high frame-skip gives a form of temporal abstraction, which simplifies the credit assignment \cite{McGovern_1998_MacroActions} and exploration \cite{Dabney_2021_TemporallyExtended} problems of RL \cite{kalyanakrishnan2021analysis}.

The results in \reftable{tab:unseen-opponents} indeed confirm that strategies learned in the training runs against Ryu tend to generalize poorly to different opponent characters---especially as their difficulty level increases. Subsequent finetuning can improve this, but only to a limited degree (\reftable{tab:finetuning-results}). For reference: a \textit{Separated (4-16,32)} agent trained from scratch for a full period of 10 million steps against \textit{any} of the opponent characters individually, can obtain a 100\% winrate against that character. This suggests that the built-in AI for \textit{all} characters in the game may be susceptible to exploitative strategies, but these strategies could be different for different characters. \reffigure{fig:combo_seq} confirms that the agent tends to overwhelmingly prefer a single combo move at each point in time in the training process, but which combo move it prefers changes as the opponent character we train against changes. Similar behavior was found under the \textit{Single Finetuning} regime (plots omitted due to space constraints).

\section{Conclusion}

This paper investigated using reinforcement learning (RL) to not only decide which actions (types of movements and attacks) to use in fighting games, but also to decide---conditioned on the game state---which frame-skip value to use. We trained agents using proximal policy optimization \cite{schulman2017proximal} with neural network architectures modified to include policy outputs for frame-skip value selection. This was done in the \textit{Street Fighter II - Special Champion Edition} game, using the FightLadder \cite{Li_2024_FightLadder} framework. Both training and evaluation was done against various opponent characters being controlled by the game's built-in AI.

We found that RL can be used to effectively select frame-skip values autonomously (rather than using predetermined fixed values), but in our setup, agents tended to learn to pick rather high frame-skip values. Agents using high frame-skip values as fixed, predetermined settings (not state-conditioned and autonomously selected) worked similarly well. It turned out that all built-in AIs in this game are susceptible to exploitative strategies that repeatedly execute the same action or small subset of actions, and high frame-skip values create a form of temporal abstraction that makes it easier to learn such exploitative strategies. 

These findings lead to the insight that it is important to be careful with evaluations based on built-in AIs in these games. Future work---both general RL work in fighting games, as well as specifically work focused on learning to select frame-skip values as a part of the policy---ought to include different agents, such as trained ones, for more robust evaluations. Training exploiters with seemingly excessively high frame-skip values could serve as a good, cheap test of robustness or exploitability of agents. 

We still believe that further research into learning to select action durations could lead to more interesting behaviors when training and evaluating against more robust opponents in future work. As we found in this paper, high frame-skip values can bring an advantage in the form of temporal abstraction, but lower frame-skip values may---depending on state---become necessary for adequate response times against stronger opponents. Adding constraints around the frame-skip values that agents are allowed to select could be an interesting direction for such future research. For example, agents could be constrained to a maximum number of decisions per full game, or per unit of time (e.g., second), and hence required to intelligently distribute their decisions in a possibly non-uniform manner according to those constraints. Constraints could be motivated by compute time or limits in reaction times and actions per second that human players have. Such work would be closely aligned with recently proposed research directions on RL agents that do not solely optimize for their performance in a game or environment, but also for the computation cost used to do so \cite{Orenstein_2025_Towards}.

\begin{credits}

\subsubsection{\discintname}
The authors have no competing interests to declare that are relevant to the content of this article.
\end{credits}
%
%
%
\bibliographystyle{splncs04}
\bibliography{ref,Dennis-Soemers-Bib}
\end{document}